\documentclass[10pt,twocolumn,letterpaper]{article}
\usepackage[pagenumbers]{cvpr}
\usepackage{microtype}
\usepackage{mathtools}
\usepackage{array}
\usepackage{multirow}
\usepackage{colortbl}

\definecolor{cvprblue}{rgb}{0.21,0.49,0.74}
\usepackage[pagebackref,breaklinks,colorlinks,allcolors=cvprblue]{hyperref}

\def\confName{CVPR}
\def\confYear{2026}
\title{A2I-VLA: A Two-Stage Vision–Language–Action model with Semantic–Physical Alignment and Integration}

\author{%
	Kun Dai\textsuperscript{1}, Tianqi Wang\textsuperscript{1}, Haibo Lei\textsuperscript{2},
	Chaoyang Li\textsuperscript{1}, Kai Zhang\textsuperscript{2}, Fuyuan Qiu\textsuperscript{1}\\
	Lijun Zhao\textsuperscript{1}, Ruifeng Li\textsuperscript{1}, Chuqing Cao\textsuperscript{3},
	Tao Xie\textsuperscript{1,*}, Wenming Tan\textsuperscript{2,*}\\[0.5em]
	{\small\textsuperscript{1} Harbin Institute of Technology}\\
	{\small\textsuperscript{2} Hikvision Research Institute}\\
	{\small\textsuperscript{3} Anhui Polytechnic University}%
}
\makeatletter
\providecommand{\@LN@col}[1]{}
\g@addto@macro\@maketitle{
\noindent\begin{minipage}{\textwidth}
	\centering
	\captionsetup{type=figure}
	\includegraphics[width=\linewidth]{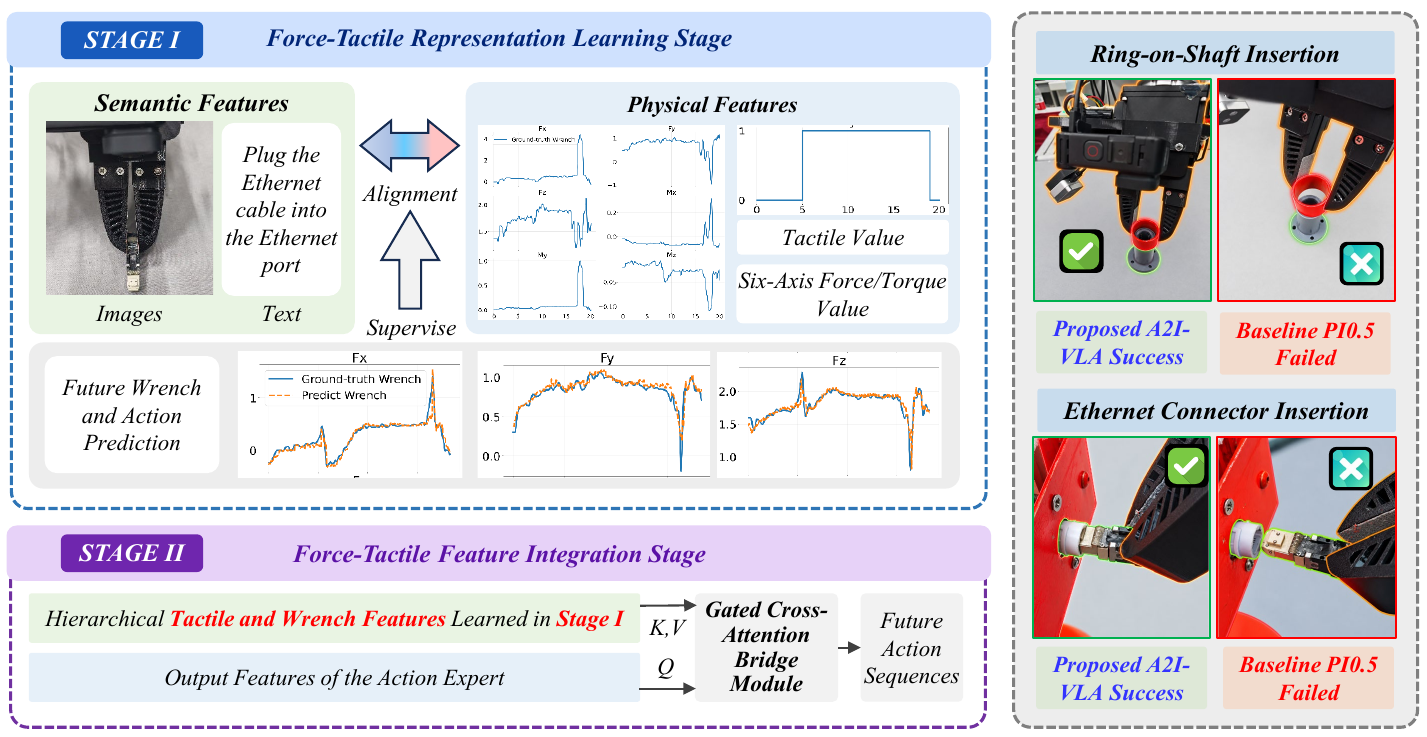}
	\caption{A2I-VLA adopts a two-stage framework to integrate wrench and tactile features into the VLA model. Stage I aims to realize semantic-physical alignment and scene-conditioned force-tactile understanding.
	Stage II integrates the force-tactile representations learned in Stage I into the VLA model to guide action generation.}
	\label{intro_fig}
	\vspace{2em}
\end{minipage}\par
}
\makeatother
\begin{document}
	\maketitle

\begin{abstract}
Force and tactile feedback provide essential information about physical interactions in contact-rich manipulation that visual observations alone cannot reliably capture.
However, integrating these signals into vision–language–action (VLA) models remains challenging owing to the representational gap between visual–language semantics and physical measurements.
To address this challenge, we propose A2I-VLA, a two-stage VLA framework that first learns semantically \textbf{A}ligned force–tactile representations and then \textbf{I}ntegrates them into a \textbf{VLA} model.
In Stage I, a cross-modal alignment loss is introduced to promote correspondence between semantic and physical representations. 
The tactile and force feature streams are then integrated through a hierarchical multimodal fusion network, and the resulting representations are jointly supervised by future wrench and action prediction losses. 
The former encourages the model to anticipate physical responses, while the latter associates force–tactile observations with task-relevant manipulation behaviors.
In Stage II, a gated cross-attention bridge module adaptively integrates the hierarchical physical features learned in Stage I into the VLA model to guide action generation.
Real-world experiments demonstrate that the proposed A2I-VLA significantly outperforms the baseline $\pi_{0.5}$ across three contact-rich manipulation tasks, improving the success rate from 70\% to 95\%.
\end{abstract}

	\section{Introduction}
\label{sec:intro}
\noindent\hspace*{1em}
Recent advances in vision-language-action (VLA) models \cite{openvla,pi0,pi05,amin2026pi,PI07,zheng2026x,lingbotvla,lingbotvla2,yang2026abot,luo2026being,ye2026starvla,yu2026twinbrainvla} have demonstrated considerable potential across diverse robotic manipulation tasks by connecting visual observations and task instructions with action generation.
However, contact-rich manipulation requires robots to perceive and respond to physical interactions that visual observations alone cannot reliably capture \cite{forcevla, forcevla2, forcedelta, tactilevla}.
In such settings, visually similar configurations may correspond to different contact conditions and require different actions.
Force and tactile feedback can help resolve this ambiguity by providing complementary information about interaction forces and local contact, making their effective integration important for contact-rich manipulation.

Existing methods \cite{forcevla,forcevla2,tla,VTLA,tactilevla} typically encode force or tactile measurements through dedicated branches and directly integrate the resulting tokens with visual–language representations from pretrained backbones \cite{beyer2024paligemma,Qwen2,team2024gemma}.
For example, Tactile-VLA \cite{tactilevla} introduces tactile tokens into the VLM input sequences, while ForceVLA \cite{forcevla} fuses force tokens with VLM outputs through a force-aware Mixture-of-Experts module.
Experimental results of these methods demonstrate the benefits of incorporating physical feedback into action generation.
However, a challenge lies in the heterogeneous information encoded by these modalities.
Force measurements capture interaction forces and torques, while tactile signals characterize local contact state. 
In contrast, visual features encode object appearance and spatial relationships, whereas language features convey task instructions and goals.
Although these modalities provide complementary information for manipulation, the force–tactile features capture physical interactions, whereas the visual–language features encode semantic information.
Existing direct integration frameworks \cite{forcevla,forcevla2,VTLA,tactilevla} overlook the fundamental differences between these heterogeneous representations, introducing conflicts during training.
As a result, force–tactile feedback cannot be effectively leveraged to conduct contact-rich manipulation.

To address this challenge, we propose A2I-VLA, a two-stage framework that first aligns semantic and physical representations and then integrates the learned physical features into VLA models for action generation, as illustrated in Fig. \ref{intro_fig}.
In the \emph{force-tactile representation learning stage}, separate semantic and physical branches encode visual–language observations and current wrench and tactile measurements, respectively.
A cross-modal alignment loss promotes semantic-physical correspondence between the two streams.
The semantic and physical features are then integrated through a hierarchical multimodal fusion network, producing multi-level physical features enriched with visual–language context.
These features are subsequently concatenated to form a unified representation for future wrench and action prediction.
The former encourages the model to anticipate physical responses, while the latter links force–tactile observations to task-relevant manipulation behaviors.
Together, the alignment and prediction losses guide the learning of physical representations.

In the \emph{force-tactile feature integration stage}, A2I-VLA transfers the learned feature extractor and retains the hierarchical tactile and wrench features. 
A gated cross-attention bridge module utilizes the action expert’s output features as queries to retrieve the two types of physical context through separate cross-attention branches.
Learned weighting factors modulate the retrieved features before adding them to action features through residual connections.
This design enables A2I-VLA to adjust the contribution of physical context during action generation.
During training, a two-phase strategy is adopted to progressively optimize the overall architecture. 
Specifically, A2I-VLA first trains only the bridge module while keeping all other components frozen.
Subsequently, all components are jointly fine-tuned.

Our contributions are summarized as follows:
\begin{itemize}
	\item We propose A2I-VLA, a two-stage VLA model for contact-rich manipulation that incorporates force and tactile feedback to guide action generation.
	\item We develop a two-stage framework for force–tactile feature alignment and integration into VLA models. 
	In the first stage, A2I-VLA integrates semantic–physical feature alignment with future wrench and action prediction to guide the learning of hierarchical force–tactile features.
	In the second stage, A2I-VLA integrates physical features learned in the first stage into the VLA through a gated cross-attention bridge module.
	\item We evaluate A2I-VLA on ring-on-shaft insertion, blackboard wiping, and ethernet connector insertion tasks. Compared with $\pi_{0.5}$, A2I-VLA improves the task success rate from 70.00\% to 95.00\%.
\end{itemize}

	\section{Related Work}
\label{sec:related}
\subsection{Vision-Language-Action Models}
\noindent\hspace*{1em}
Vision-language-action (VLA) models \cite{openvla,pi0,pi05,amin2026pi,zheng2026x,lingbotvla,lingbotvla2,yang2026abot,luo2026being,ye2026starvla,PI07} integrate visual perception, language understanding, and action prediction within a unified end-to-end framework.
OpenVLA \cite{openvla} formulates robot control as the autoregressive generation of discretized action tokens and fine-tunes a pretrained vision–language model \cite{touvron2023llama} using a next-token prediction objective.
Unlike OpenVLA, $\pi_{0}$ \cite{pi0} combines a pretrained vision–language backbone \cite{beyer2024paligemma} with a flow-matching technique to generate continuous action chunks.
This continuous action generation paradigm is widely adopted in subsequent VLA studies.
Building upon $\pi_{0}$, $\pi_{0.5}$ \cite{pi05} introduces hierarchical subtask inference to enable long-horizon manipulation in unseen environments.
As reinforcement learning continues to improve performance on dexterous manipulation tasks, $\pi_{0.6}$ \cite{amin2026pi} extends flow-based continuous action generation with RECAP, a reinforcement learning framework that improves policies through real-world deployment.
To improve cross-embodiment generalization, X-VLA \cite{zheng2026x} employs embodiment-specific soft prompts to accommodate diverse robotic embodiments.
LingBot-VLA series \cite{lingbotvla,lingbotvla2} integrates vision-language understanding and continuous action generation through a Mixture-of-Transformers architecture, with depth-feature distillation to enhance spatial perception.
ABot-M0 \cite{yang2026abot} combines unified pretraining on heterogeneous robot data with action manifold learning, predicting action sequences rather than velocity targets.

\subsection{Physical Modalities in VLA Models}
\noindent\hspace*{1em}
Force and tactile feedback play essential roles in contact-rich robotic manipulation, providing physical interaction information that cannot be reliably inferred from visual observations alone.
Recent studies \cite{forcevla, forcevla2, forcedelta, phaforce, tla, VTLA, tactilevla, lou2026dream, zhang2026unitacvla} have explored integrating these modalities into the VLA models.
For force-based methods, ForceVLA \cite{forcevla} introduces a force-aware Mixture-of-Experts (MoE) module to adaptively fuse six-axis force-torque measurements with visual-language features for contact-aware action generation.
ForceVLA2 \cite{forcevla2} further incorporates force-based prompts into task reasoning and employs a MoE module to enable force-position control.
ForceDelta-VLA \cite{forcedelta} adopts a distillation framework that utilizes a lightweight policy to refine reference actions.
PhaForce \cite{phaforce} introduces a slow-fast policy that coordinates visual-force planning and high-frequency residual corrections according to predicted contact states and task phases.
For tactile-based methods, 
TLA \cite{tla} and VTLA \cite{VTLA} incorporate sequential tactile observations into the encoder of LM or VLM to predict actions.
Tactile-VLA \cite{tactilevla} combines tactile feedback with hybrid position-force control to enable language-guided physical interaction.
AT-VLA \cite{atvla} introduces adaptive tactile injection and a dual-stream strategy to selectively incorporate tactile feedback while preserving pretrained VLA capabilities.

However, most existing methods adopt input-level integration without explicitly addressing the representational gap between force–tactile measurements and visual–language semantics.
This gap introduces conflicts between heterogeneous features, limiting the effective use of physical feedback in contact-rich manipulation.

	\section{Method}
\label{sec:method}

\subsection{Overall Architecture of A2I-VLA}
\label{sec:overview}
\begin{figure*}[t]
	\centering
	\includegraphics[width=
	\textwidth]{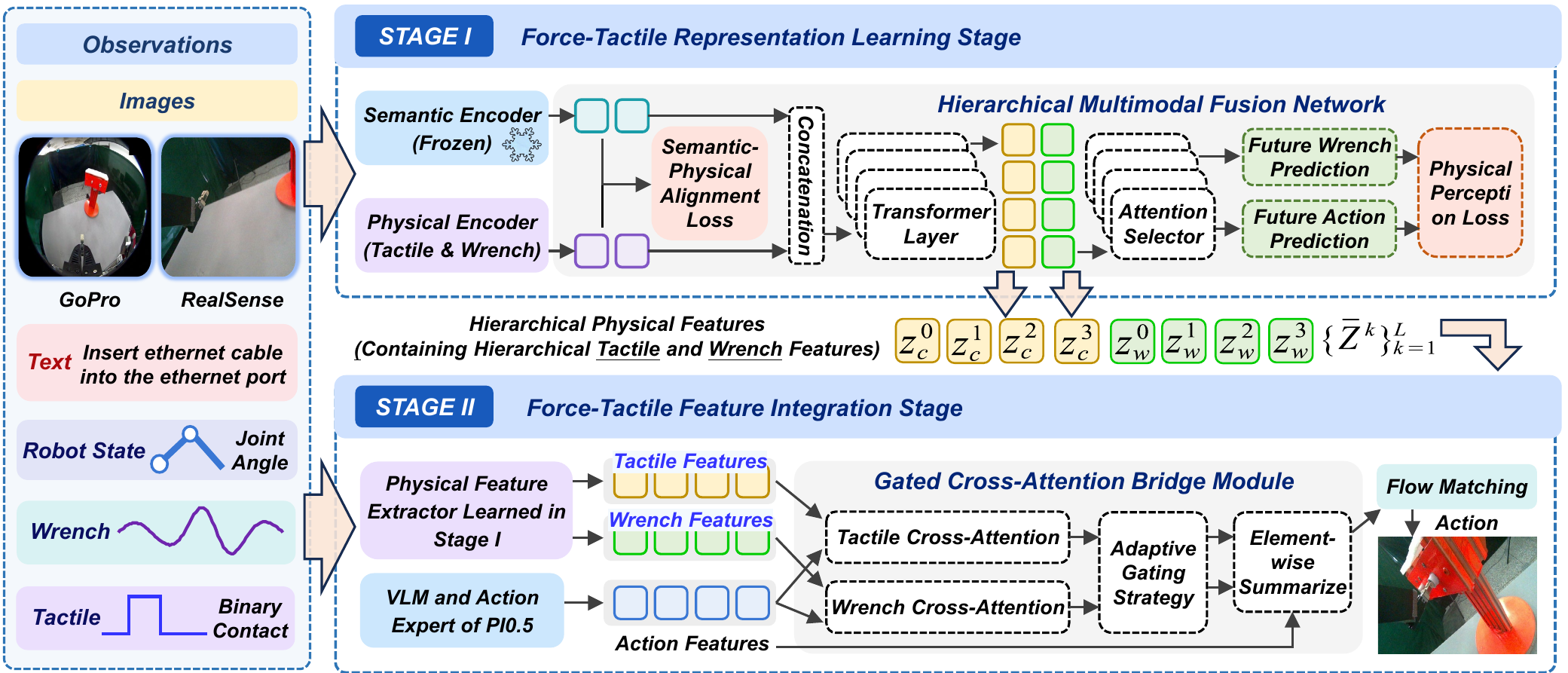}
	\caption{Architecture of the proposed A2I-VLA. 
	In Stage I, A2I-VLA aligns semantic and physical features through a semantic–physical alignment loss. In addition, future wrench and action prediction tasks are introduced to guide the learning of task-aware force–tactile representations.
	In Stage II, A2I-VLA integrates the hierarchical tactile and wrench features learned in the first stage into the VLA framework through a gated cross-attention bridge module.}
	\label{fig:overview}
\end{figure*}

\noindent\hspace*{1em}
As illustrated in Fig.~\ref{fig:overview}, A2I-VLA consists of a \emph{force-tactile representation learning stage} and a \emph{force-tactile feature integration stage}.
Stage I realizes the semantic-physical alignment and scene-conditioned force-tactile understanding.
Stage II integrates the hierarchical physical features learned in Stage I into the VLA model.

At time $t$, A2I-VLA receives two RGB images $I_t=\{I_g,I_r\}$, a language instruction $\ell$, robot state $p_t$, a current wrench value $w_t \in \mathbb{R}^{6}$, and a current tactile value $c_t \in \{0,1\}$.
The wrench $w_t$ contains three force and three torque components, while $c_t$ is a binary scalar taking values of 0 or 1.
Given these inputs, the policy predicts an action chunk $A_t=[a_t,\ldots,a_{t+H_a-1}]$.

In Stage I, a physical encoder extracts physical features $F_t^{phy}$ from $w_t$ and $c_t$, while a parallel semantic branch employs pretrained VLA components to extract semantic features $F_t^{sem}$ from $I_t$, $\ell$, and $p_t$.
A cross-modal alignment loss $\mathcal{L}_{align}$ promotes correspondence between the two streams before fusion.
A hierarchical multimodal fusion network then jointly processes the semantic and physical features, producing multi-level physical features $\{\bar Z^k\}_{k=1}^{L}$, where $L$ denotes the number of fusion layers.
At each layer, an attention selector processes the $\bar Z^k$ into a descriptor $\bar H^k$.
Future wrench prediction loss $\mathcal{L}_w$ guides the model to anticipate physical responses, while action prediction loss $\mathcal{L}_a$ associates the learned features with task-relevant behaviors.

In Stage II, the learned physical encoder and hierarchical multimodal fusion network form a feature extractor $\mathcal{E}_{\phi}$.
Separate projectors map the tactile and wrench features from all fusion layers into $F_c$ and $F_w$.
A gated cross-attention bridge uses the action expert's output as queries and these context features as keys and values to produce action features $\widetilde F_{act}$.
The action output projection then predicts flow vectors for action generation.

\subsection{Force-Tactile Representation Learning Stage}
\label{sec:alignment}
\noindent\hspace*{1em}
Stage I aims to achieve semantic–physical alignment and force–tactile understanding.

\textbf{Parallel Semantic and Physical Encoder.}
As shown in Fig. \ref{fig:overview}, the semantic encoder directly uses the pretrained image encoder of $\pi_{0.5}$ to extract visual features from two RGB images $I_t=\{I_g,I_r\}$.
The language instruction $\ell$ and robot state $p_t$ are jointly tokenized using the $\pi_{0.5}$ tokenizer and embedded through the pretrained text embedding layer.
The visual and language features are concatenated to form semantic features $F_t^{sem}$.
These pretrained components remain frozen during the Stage I.

The physical encoder uses MLPs to independently encode tactile and wrench measurements, whose outputs are then concatenated to form the physical features $F_t^{phy}$.

\textbf{Hierarchical Multimodal Fusion Network.} Learnable modality embeddings are added to $F_t^{sem}$ and $F_t^{phy}$ to distinguish the semantic and physical streams. 
The resulting features are denoted as $S_t$ and $P_t$, respectively.
Subsequently, the network first employs two separate MLPs to process $S_t$ and $P_t$, generating $U^{sem}_{t}$ and $U^{phy}_{t}$, respectively.
These features are then concatenated and handled through $L$ Transformer layers \cite{vaswani2017attention} to model their interactions, generating hierarchical physical features $\{\bar Z^k\}_{k=1}^{L}$.

At each layer, two physical features $\bar Z^k=[z_c^k,z_w^k]$ preserve their respective tactile and wrench positions while incorporating visual-language context through self-attention.
For each $\bar Z^k$, an attention selector employs an MLP to predict a score for each modality, followed by softmax to obtain attention weights.
The weighted features are summed and projected to generate a physical descriptor $\bar H^k$.

\textbf{Semantic-Physical Alignment Loss.}
In this part, we align $U^{sem}_{t}$ and $U^{phy}_{t}$ to mitigate the discrepancies between heterogeneous features.
Specifically, a shared learnable query $Q$ independently attends to semantic and physical features via the cross-attention, generating $R^{sem}_{t}$ and $R^{phy}_{t}$ for cross-modal alignment.
\begin{equation}
	\begin{gathered}
		R^{sem}_{t} =Q+\operatorname{CrossAtt}(Q,U^{sem}_{t},U^{sem}_{t}),\\
		R^{phy}_{t} =Q+\operatorname{CrossAtt}(Q,U^{phy}_{t},U^{phy}_{t})
	\end{gathered}
	\label{eq:query}
\end{equation}

The shared query $Q$ generates an equal number of embeddings from $U^{sem}_{t}$ and $U^{phy}_{t}$, regardless of their different token counts.
We minimize the $L_{2}$ distance between $R^{sem}_{t}$ and $R^{phy}_{t}$, while applying variance to avoid constant features across different observations.
Finally, the semantic-physical alignment loss $\mathcal{L}_{align}$ is defined as:
\begin{equation}
	\begin{gathered}
		\mathcal{L}_{align} = \frac{1}{K}\sum_{i=1}^{K}\left\|R^{sem}_{t}(i)-R^{phy}_{t}(i)\right\|_2+ \\
		\frac{1}{2}\bigl[\mathcal{V}(R^{sem}_{t})+\mathcal{V}(R^{phy}_{t})\bigr]
	\end{gathered}
	\label{eq:align}
\end{equation}
where $K$ means the number of tokens, $R^{sem}_{t}(i)$ and $R^{phy}_{t}(i)$ denote the $i$-th token of $R^{sem}_{t}$ and $R^{phy}_{t}$. $\mathcal{V}(\cdot)$ is defined as follow:
\begin{equation}
	\mathcal{V}(R) = \frac{1}{KC}\sum_{i=1}^{K}\sum_{j=1}^{C}
	\max\!\left(0,\,1-\sigma(R(i,j))\right)
	\label{eq:variance}
\end{equation}
where $C$ means the feature dimension and
$\sigma(R(i,j))$ denotes the variance of the $(i,j)$-th value of $R$ across all samples in the batch.

\textbf{Physical Perception Loss.}
We concatenate the hierarchical descriptors $\{\bar H^k\}_{k=1}^{L}$ and feed them into two separate MLP heads to predict future wrench sequences $\bar W_t^f$ and action sequences $\bar A_t^f$, respectively.
Wrench prediction encourages the model to anticipate physical responses, while action prediction links the learned physical features to manipulation behavior.
Therefore, the physical perception loss contains a future wrench prediction loss $\mathcal{L}_w$ and  an action prediction loss $\mathcal{L}_a$.
\begin{equation}
	\begin{gathered}
		\mathcal{L}_w= \frac{1}{K_w}\sum_{i=1}^{K_w}\left\|\bar W_t^f(i)-W_t^f(i)\right\|_2,\\
		\mathcal{L}_a=\frac{1}{K_a}\sum_{j=1}^{K_a}\left\|\bar A_t^f(j)-A_t^f(j)\right\|_2
	\end{gathered}
	\label{eq:physical_perception}
\end{equation}
where $W_t^f$ and $A_t^f$ are the ground-truth future wrench and action sequences. $K_w$ and $K_a$ mean the length of sequences.

Finally, the loss function of Stage I is defined as:
\begin{equation}
	\mathcal{L}_{stage1}=\lambda_{align}\mathcal{L}_{align}
	+\lambda_w\mathcal{L}_w+\lambda_a\mathcal{L}_a,
	\label{eq:stage1}
\end{equation}

\subsection{Force-Tactile Feature Integration Stage}
\label{sec:bridge}
\noindent\hspace*{1em}
Stage II integrates the hierarchical force-tactile features learned in Stage I into the VLA model to guide action generation.
We combine the physical encoder and hierarchical multimodal fusion network learned in Stage I as a physical feature extractor $\mathcal{E}_{\phi}$.
The attention selectors and prediction heads are discarded.

\textbf{Gated Cross-Attention Bridge Module.}
At the $k$-th fusion layer, we extract the tactile feature $z_c^k$ and wrench feature $z_w^k$ from $\bar Z^k$ according to their token positions.
These features incorporate semantic context, enabling the network to interpret similar physical measurements in accordance with the visual context.
Then, separate linear layers map $z_c^k$ and $z_w^k$ into compact features $m_c^k$ and $m_w^k$.
For $L$ fusion layer, we obtain $F_c \in \mathbb{R}^{L \times C}$ and $F_w \in \mathbb{R}^{L \times C}$

Subsequently, we employ the VLM and action expert of $\pi_{0.5}$ to process input images, language and noisy action chunk.
Let $F_{act}$ denote the final output features of the action expert.
We then utilize $F_{act}$ as queries to independently aggregate tactile and wrench features through two separate cross-attention branches.
Subsequently, a learnable gating strategy adaptively modulates the output features, which are integrated into $F_{act}$ through a residual connection to guide action generation.
This process can be formulated as follows:
\begin{equation}
	\begin{gathered}
		D_c=\operatorname{CrossAtt}(F_{act}, F_c, F_c),\\
		D_w=\operatorname{CrossAtt}(F_{act}, F_w, F_w),\\
		[G_c,G_w]=\operatorname{Sigmoid}\left(\operatorname{Linear}
		\!\left(\operatorname{Cat}(F_{act},D_c,D_w)\right)\right),\\
		\widetilde F_{act}=F_{act} + G_c\odot \operatorname{Linear}_{c}(D_c) + G_w\odot \operatorname{Linear}_{w}(D_w)
	\end{gathered}
	\label{eq:bridge}
\end{equation}

The updated action features $\widetilde F_{act}$ are processed by the flow matching technique to generate action chunk. 
Notably, the linear layers $\operatorname{Linear}_{c}$ and $\operatorname{Linear}_{w}$ are initialized to zero so that $\widetilde F_{act}=F_{act}$ at initialization.

\textbf{Loss Function of Stage II.}
Following $\pi_{0.5}$, we adopt the same loss function to minimize the mean squared error between the predicted flow vector and the target flow vector.

\textbf{Two-Phase Training Strategy for Stage II.}
We adopt a two-phase training strategy to progressively integrate the learned representations into the VLA.
Firstly, we train only the gated cross-attention bridge module while keeping the base VLA and physical feature extractor $\mathcal{E}_{\phi}$ frozen.
Subsequently, we unfreeze all modules and jointly optimize all parameters.
oth phases use the same flow-matching loss.

	\section{Experiments}
\label{sec:experiments}
\subsection{Experimental Setup}
\noindent\hspace*{1em}
\textbf{Architecture Details.}
The hierarchical multimodal fusion network consists of $L=4$ Transformer layers, each with a feature dimension of $d=256$ and eight attention heads. The semantic–physical alignment loss employs $K=4$ shared queries, with loss weights set to $\lambda_{align}=0.1$ and $\lambda_w=\lambda_a=1$.

\textbf{Training Strategy.}
We train the A2I-VLA on eight A100 GPUs using the AdamW optimizer for both stages.
In Stage I, we freeze the semantic encoder pretrained by the $\pi_{0.5}$ and train $1 \times 10^{4}$ iterations with a batch size of 512 and a weight decay of $1 \times 10^{-4}$. 
The learning rate is gradually increased to $1 \times 10^{-4}$ over the first 1000 warmup iterations, followed by the cosine decay strategy.
During Stage II, A2I-VLA is trained for $5 \times 10^{4}$ iterations with a batch size of 64. 
During the first 1000 iterations, only the gated cross-attention bridge module is trained, with the learning rate increasing from $0$ to $5 \times 10^{-5}$. 
The base VLA and feature extractor are then unfrozen for an additional 1000 iteration warmup.
Afterward, the learning rate is maintained at $5 \times 10^{-5}$.

\begin{figure}[t]
	\centering
	\includegraphics[width=\linewidth]{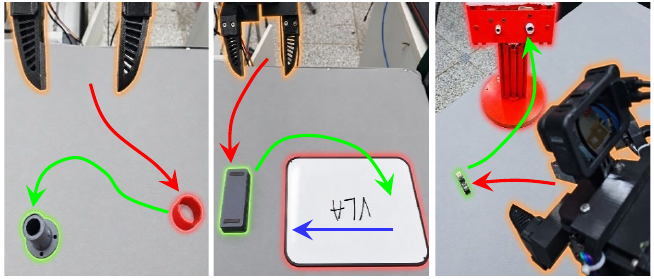}
	\caption{We conduct experiments on ring-on-shaft insertion, blackboard wiping, and ethernet connector insertion tasks. The diameter of the shaft and inner diameter of the ring are 32.0mm and 34.5mm.}
	\label{fig:tasks}
\end{figure}

\begin{figure}[t]
	\centering
	\centering
	\includegraphics[width=0.88\linewidth]{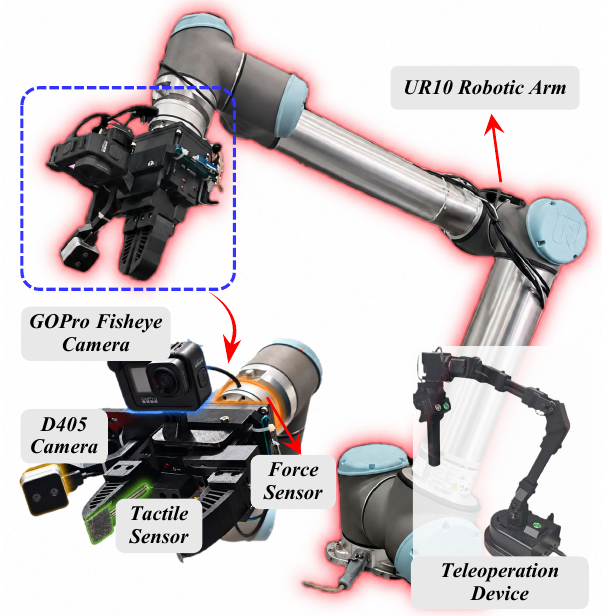}
	\caption{The experimental platform that contains a UR10 robotic arm, a GoPro fisheye camera, an Intel RealSense D405 camera, a six-axis force/torque sensor, and two tactile sensors.}
	\label{platform}
	\vspace{-3mm}
\end{figure}

\begin{table*}[t]
	\centering\small
	\caption{Comparison of best-checkpoint success rates (\%) across three real-world manipulation tasks. The highest success rate among the final five evaluated checkpoints is reported. Each checkpoint is evaluated over 20 real-robot trials.}
	\label{Table_Main_Exp}
	\setlength{\tabcolsep}{6pt}
	\renewcommand{\arraystretch}{1.1}
	\begin{tabular}{lcccc}
		\toprule[1.25pt]
		Method & Ring-on-Shaft & Blackboard Wiping & Ethernet Insertion & Average \\
		\midrule
		$\pi_{0.5}$ \cite{pi05} (baseline) & 17/20 (85.00\%) & 16/20 (80.00\%) & 9/20 (45.00\%) & 42/60 (70.00\%) \\
		ForceVLA \cite{forcevla} & 18/20 (90.00\%) & 16/20 (80.00\%) & 10/20 (50.00\%) & 44/60 (73.33\%) \\
		\cellcolor{gray!10}A2I-VLA (ours) & \cellcolor{gray!10}\textbf{20/20 (100.00\%)} & \cellcolor{gray!10}\textbf{20/20 (100.00\%)} & \cellcolor{gray!10}\textbf{17/20 (85.00\%)} & \cellcolor{gray!10}\textbf{57/60 (95.00\%)} \\
		\bottomrule[1.25pt]
	\end{tabular}
\end{table*}


\begin{figure*}[t]
	\centering
	\includegraphics[width=\linewidth]{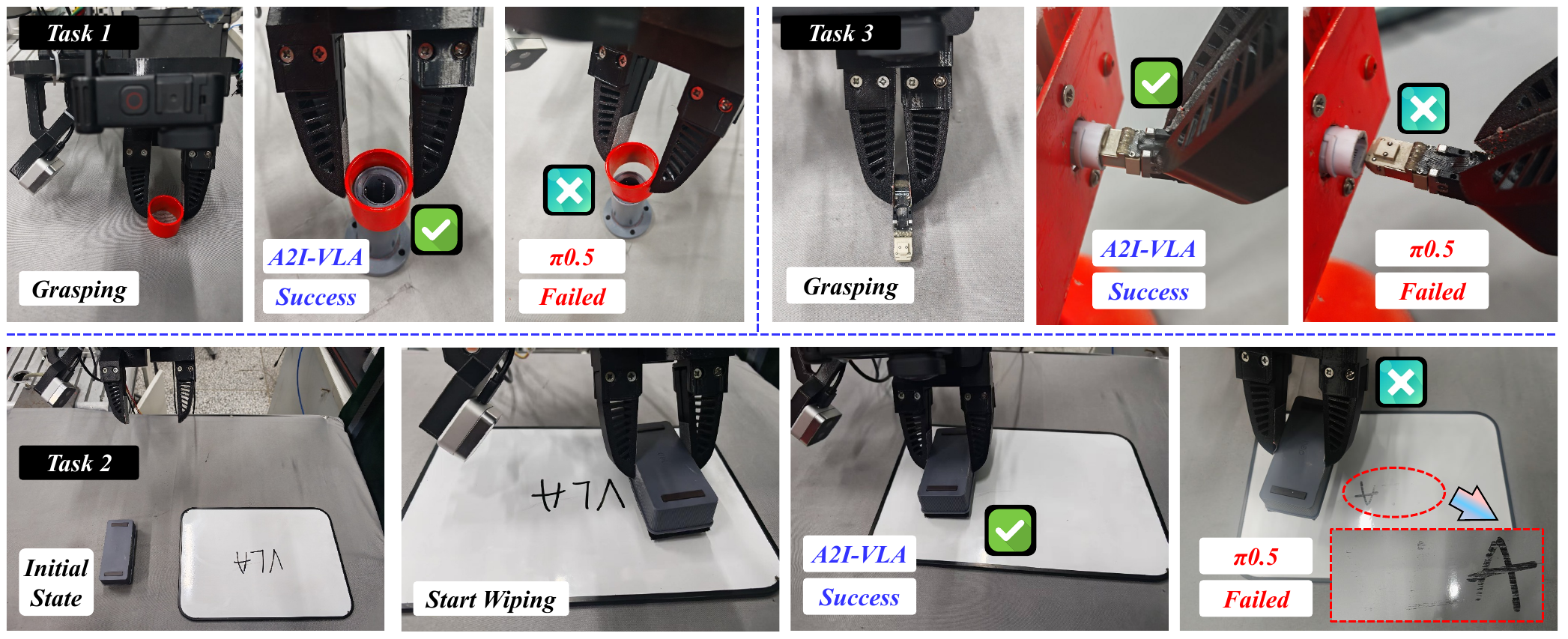}
	\caption{Qualitative comparison between the proposed A2I-VLA and the baseline $\pi_{0.5}$ across three contact-rich manipulation tasks. A2I-VLA consistently outperforms the baseline across all three tasks.}
	\label{task_execute_comparison}
\end{figure*}

\textbf{Data Augmentation.}
During Stage II, we apply random brightness, contrast, and saturation jitter to input images. 
Additionally, we augment the raw force components $(F_x,F_y,F_z)$ before normalization. 
For each sample, we randomly apply one of three force augmentation strategies. 
(1) Keeping the original force measurements unchanged with a probability of $0.3$; 
(2) Adding uniform noise within $\pm0.25N$ for $F_x$ and $\pm0.4N$ for $F_y$ and $F_z$ with a probability of $0.5$; 
(3) Adding stronger uniform noise within $\pm0.5N$ for $F_x$ and $\pm0.8N$ for $F_y$ and $F_z$ with a probability of $0.2$.

\textbf{Evaluation Metrics.}
As shown in Fig. \ref{fig:tasks}, we evaluate the A2I-VLA on three real-world contact-rich manipulation tasks: ring-on-shaft insertion, blackboard wiping, and ethernet connector insertion.
The ring, eraser, and ethernet connector are placed at random positions.
For each task, we evaluate the final five saved checkpoints of each model over 20 trials per checkpoint. 
After that, we report the highest success rate achieved among the five evaluated checkpoints.



\textbf{Experimental Platform.}
As shown in Fig. \ref{platform}, the experimental platform consists of a UR10 robotic arm equipped with a GoPro camera and an Intel RealSense D405 camera at the end effector, providing a global view of the workspace and a close-up view of the contact region, respectively.
A six-axis force/torque sensor is installed on the end flange of the UR10. 
A tactile sensor mounted on each gripper provides tactile feedback via an STM32 microcontroller.
The training data is collected using the teleoperation device. 

\begin{figure*}[t!]
	\centering
	\includegraphics[width=\linewidth]{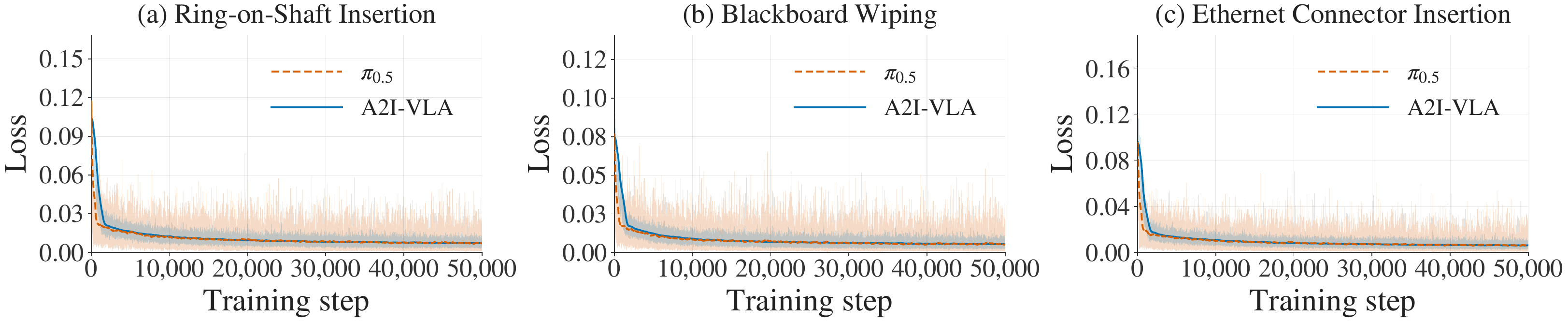}
	\caption{Stage II training loss curves for (a) ring-on-shaft insertion, (b) blackboard wiping, and (c) Ethernet connector insertion tasks. All three tasks exhibit a rapid initial loss decrease followed by gradual convergence.}
	\label{loss_comparison_three_tasks}
\end{figure*}

\begin{figure}[t!]
	\centering
	\includegraphics[width=\linewidth]{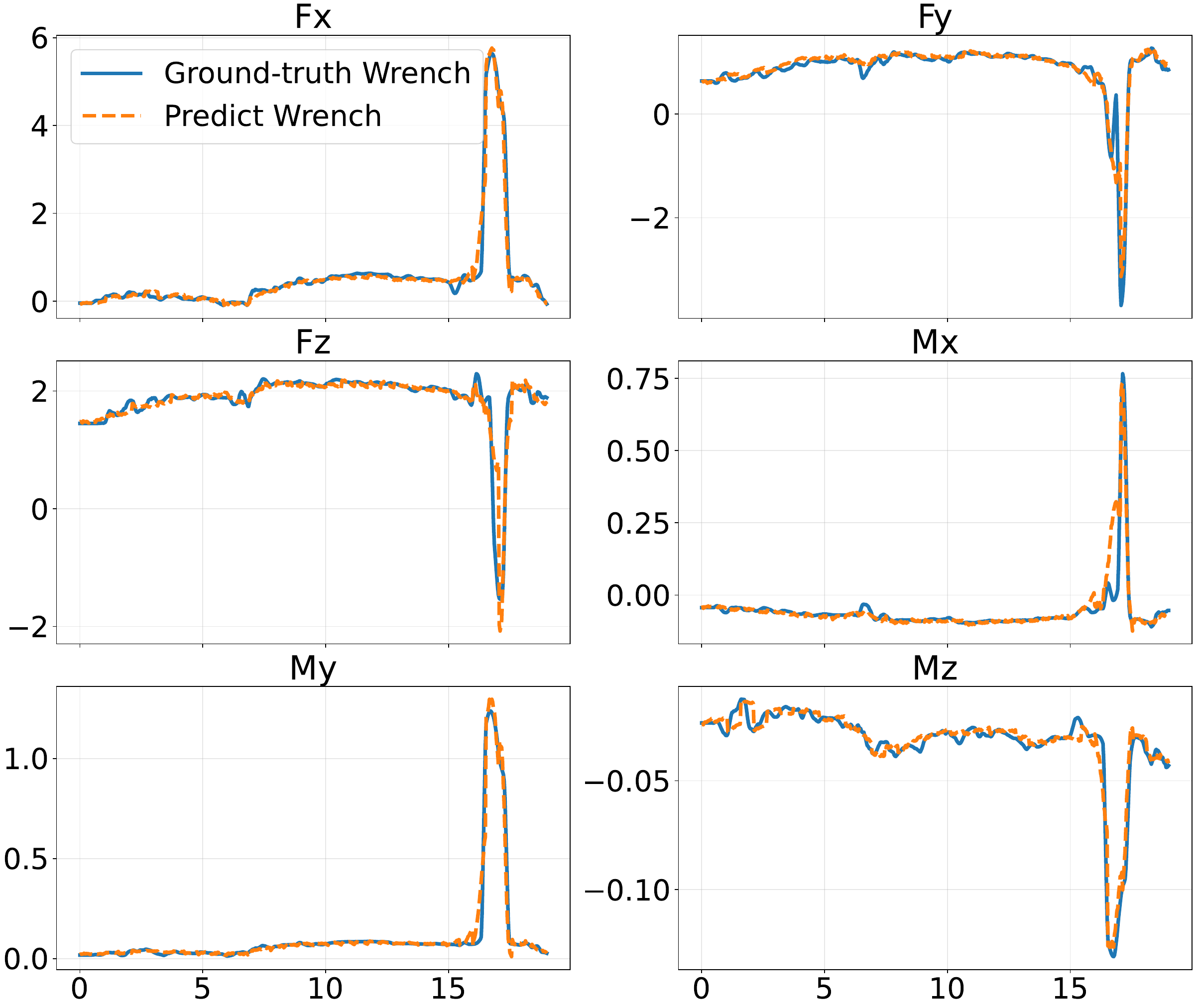}
	\caption{Future wrench prediction in Stage I. We visualize the predicted and ground-truth force and torque sequences for an example of the ethernet connector insertion task.}
	\label{predict_forceV2}
\end{figure}
\subsection{Comparison with other Methods}
\noindent\hspace*{1em}
Table \ref{Table_Main_Exp} presents a comparison of A2I-VLA with $\pi_{0.5}$ \cite{pi05} and ForceVLA \cite{forcevla} across three real-world contact-rich manipulation tasks.
Overall, A2I-VLA consistently outperforms these methods on all three tasks, achieving an average success rate of 95.00\%, compared with 70.00\% for $\pi_{0.5}$ and 73.33\% for ForceVLA.

For the ring-on-shaft insertion task, A2I-VLA achieves a 100.00\% success rate across all 20 trials, outperforming $\pi_{0.5}$ and ForceVLA, which achieve success rates of 85.00\% and 90.00\%, respectively.
For the blackboard wiping task, A2I-VLA also achieves 100.00\%, exceeding both techniques by 20.00\%.
The most significant improvements are observed in the challenging ethernet connector insertion task.
A2I-VLA achieves a success rate of 85.00\%, outperforming $\pi_{0.5}$ and ForceVLA by 40.00\% and 35.00\%, respectively.
Overall, the improvements across all three tasks demonstrate that force–tactile feedback provides complementary physical information beyond visual observations, enabling more accurate and reliable action generation in contact-rich manipulation.



As shown in Fig. \ref{task_execute_comparison}, we conduct a qualitative comparison between the proposed A2I-VLA and the baseline $\pi_{0.5}$.
For ring-on-shaft insertion and ethernet connector insertion tasks, $\pi_{0.5}$ struggles to precisely align the manipulated objects with target positions, resulting in failed insertions.
In contrast, A2I-VLA incorporates force–tactile feedback to provide additional information about physical interactions, realizing more accurate alignment. 
For the blackboard wiping task, $\pi_{0.5}$ lacks force feedback and struggles to determine whether sufficient contact force is applied between the eraser and the blackboard, resulting in incomplete removal of the writing.
In contrast, A2I-VLA incorporates force–tactile feedback to better perceive interaction forces during wiping.

\begin{table}[t!]
	\centering\small
	\caption{Training time and inference speed comparison between $\pi_{0.5}$ and A2I-VLA.}
	\label{TIME_SPEED}
	\setlength{\tabcolsep}{6pt}
	\renewcommand{\arraystretch}{1.1}
	\begin{tabular}{lcc}
		\toprule[1.25pt]
		Method & $\pi_{0.5}$ (baseline) & A2I-VLA (ours) \\
		\midrule
		Training time & 32.52h  & 44.23h (6.41h + 37.82h) \\
		Inference speed & 386.82ms  & 396.63ms \\
		\bottomrule[1.25pt]
	\end{tabular}
\end{table}

\subsection{Understanding of A2I-VLA}
\noindent\hspace*{1em}
\textbf{Training Stability Analysis.}
Fig. \ref{loss_comparison_three_tasks} compares the Stage II training loss curves of A2I-VLA and the baseline $\pi_{0.5}$.
Although $\pi_{0.5}$ exhibits a faster initial decrease in loss, the smoothed loss curves of both methods gradually converge to comparable levels as training progresses.
Besides, A2I-VLA exhibits fewer pronounced loss spikes and smaller fluctuations during subsequent training.
This phenomenon is consistently observed across all three tasks.

\textbf{Visualization of Stage I Wrench Predictions.}
To evaluate the physical prediction capability learned in Stage I, Fig.~\ref{predict_forceV2} compares the predicted and ground-truth wrench sequences during an ethernet connector insertion task. The predicted sequences closely follow the overall trends of all six force–torque components. 
In particular, the model successfully captures the positive peaks in $F_x$, $M_x$, and $M_y$, as well as the negative peaks in $F_y$, $F_z$, and $M_z$. 
This phenomenon demonstrates that Stage I learns reliable features of interaction forces.

\begin{table*}[t!]
	\centering \small
	\caption{Performance comparison between the baseline $\pi_{0.5}$ and the proposed A2I-VLA across different training iterations (i.e., 30k, 35k, 40k, 45k, and 50k) on three manipulation tasks. Average denotes the mean success rate across all three tasks and five checkpoints.}
	\label{Table_Train_Compare}
	\setlength{\tabcolsep}{3.5pt}
	\renewcommand{\arraystretch}{1.2}
	\begin{tabular}{l*{15}{c}c}
		\toprule[1.25pt]
		\multirow{2}{*}{Method}
		& \multicolumn{5}{c}{Ring-on-Shaft Insertion}
		& \multicolumn{5}{c}{Blackboard Wiping}
		& \multicolumn{5}{c}{Ethernet Insertion}
		& \multirow{2}{*}{Average} \\
		\cmidrule(lr){2-6}
		\cmidrule(lr){7-11}
		\cmidrule(lr){12-16}
		& 30k & 35k & 40k & 45k & 50k
		& 30k & 35k & 40k & 45k & 50k
		& 30k & 35k & 40k & 45k & 50k & \\
		\midrule
		$\pi_{0.5}$ (baseline)
		& \textbf{85\%} & 80\% & 70\% & \textbf{85\%} & 80\%
		& 55\% & 75\% & 10\% & \textbf{80\%} & 50\%
		& 25\% & \textbf{45\%} & 20\% & 35\% & 25\%
		& 54.67\% \\
		
		A2I-VLA (ours)
		& \textbf{100\%} & \textbf{100\%} & \textbf{100\%}
		& \textbf{100\%} & \textbf{100\%}
		& 95\% & 85\% & 90\% & \textbf{100\%} & 45\%
		& 40\% & \textbf{85\%} & 65\% & 40\% & 35\%
		& \textbf{78.67\%} \\
		\bottomrule[1.25pt]
	\end{tabular}
\end{table*}

\begin{table*}[t!]
\centering\small
\caption{Effectiveness analysis of the semantic-physical alignment operation.}
\label{Ablation_alignment}
\setlength{\tabcolsep}{6pt}
\renewcommand{\arraystretch}{1.1}
\begin{tabular}{lcccc}
	\toprule[1.25pt]
	Method & Ring-on-Shaft & Blackboard Wiping
	& Ethernet Insertion & Average \\
	\midrule
	single-stage direct concatenation
	& 16/20 (80.00\%) & 16/20 (80.00\%) & 10/20 (50.00\%) & 42/60 (70.00\%) \\
	w/o semantic-physical alignment loss
	& 18/20 (90.00\%) & 19/20 (95.00\%) & 14/20 (70.00\%) & 51/60 (85.00\%) \\
	A2I-VLA (ours)
	& \textbf{20/20 (100.00\%)}
	& \textbf{20/20 (100.00\%)}
	& \textbf{17/20 (85.00\%)}
	& \textbf{57/60 (95.00\%)} \\
	\bottomrule[1.25pt]
\end{tabular}
\end{table*}

\begin{table*}[t!]
\centering\small
\caption{Ablation studies for Stage I and II of the A2I-VLA.}
\label{Ablation}
\setlength{\tabcolsep}{6pt}
\renewcommand{\arraystretch}{1.1}
\begin{tabular}{lcccc}
	\toprule[1.25pt]
	Method & Ring-on-Shaft & Blackboard Wiping & Ethernet Insertion & Average \\
	\midrule
	w/o future action prediction (Stage I) & 19/20 (95.00\%)  & \textbf{20/20 (100.00\%)} & 15/20 (75.00\%)  & 54/60 (90.00\%) \\
	w/o force augmentation (Stage II) & 18/20 (90.00\%)  & 19/20 (95.00\%) & 13/20 (65.00\%)  & 50/60 (83.33\%) \\
	A2I-VLA & \textbf{20/20 (100.00\%)} & \textbf{20/20 (100.00\%)} & \textbf{17/20 (85.00\%)} & \textbf{57/60 (95.00\%)} \\
	\bottomrule[1.25pt]
\end{tabular}
\end{table*}

\textbf{Training Time and Inference Speed Comparison.}
In this part, we compare the training time and inference speed of A2I-VLA with those of the baseline $\pi_{0.5}$.
As shown in Table \ref{TIME_SPEED}, A2I-VLA requires 44.23 hours of training, comprising 6.41 hours for Stage I and 37.82 hours for Stage II.
In comparison, the baseline $\pi_{0.5}$ trains faster, requiring only 32.52 hours.
During inference, A2I-VLA takes 396.63ms, compared with 386.82ms for $\pi_{0.5}$, introducing an increase of 9.81ms.

\textbf{Performance Comparison Across Different Training Iterations.}
To evaluate the consistency of A2I-VLA across training checkpoints, we compare its task success rates with those of $\pi_{0.5}$ at different training iterations.
As shown in Table~\ref{Table_Train_Compare}, A2I-VLA outperforms $\pi_{0.5}$ in 14 out of 15 comparisons.
Across all five checkpoints, A2I-VLA achieves an average success rate of 78.67\%, compared with 54.67\% for $\pi_{0.5}$, representing an improvement of 24.00\%.
These results are consistent with the best-checkpoint comparison in Table \ref{Table_Main_Exp}, demonstrating that the performance gains are not limited to the best-performing checkpoints.
Nevertheless, performance fluctuations remain across training iterations, particularly in the blackboard wiping and ethernet connector insertion tasks.

\subsection{Ablation Studies}
\noindent\hspace*{1em}
\textbf{Analysis of Semantic–Physical Alignment.}
To evaluate the effectiveness of the proposed semantic-physical alignment, we conduct two comparative experiments. 
Firstly, following the Tactile-VLA \cite{tactilevla}, we replace the two-stage framework with a single-stage architecture, where wrench and tactile tokens are directly concatenated with visual and language tokens. 
Secondly, we remove the semantic-physical alignment loss from Stage I and retrain A2I-VLA from scratch.

As shown in Table~\ref{Ablation_alignment}, the single-stage direct concatenation method achieves an average success rate of 70.00\%, which is 25.00\% lower than 95.00\%. 
The performance difference is particularly pronounced in the ethernet connector insertion task, where the success rate increases from 50.00\% to 85.00\%. 
Furthermore, removing the semantic-physical alignment loss reduces the average success rate to 85.00\%. 
These results demonstrate the effectiveness of semantic-physical alignment in improving the integration of heterogeneous representations for contact-rich manipulation.

\textbf{Analysis of Different Modules.}
We further conduct ablation experiments to investigate the effectiveness of different modules. 
As shown in Table \ref{Ablation}, removing future action prediction reduces the average success rate from 95.00\% to 90.00\%.
Moreover, removing the force augmentation strategy decreases the average success rate to 83.33\%. 
This result suggests that force augmentation helps mitigate the effects of measurement fluctuations in six-axis force/torque sensor, thereby improving the robustness to variations in force measurements.

	\section{Conclusion}
\noindent\hspace*{1em}
In this paper, we introduce A2I-VLA, a two-stage VLA framework that integrates force–tactile feedback to enable more accurate contact-rich manipulation.
In Stage I, semantic–physical alignment, hierarchical multimodal fusion, and future wrench-action prediction are employed to learn task-aware force–tactile representations. 
Building upon the representations learned in Stage I, Stage II integrates hierarchical tactile and wrench features into the VLA model through a gated cross-attention bridge module, enabling physical context to adaptively guide action generation.
We evaluate A2I-VLA on three real-world manipulation tasks, including ring-on-shaft insertion, blackboard wiping, and ethernet connector insertion. 
Experimental results demonstrate that A2I-VLA consistently outperforms the baseline across all three tasks.

	{\small
		\bibliographystyle{ieeenat_fullname}
		\bibliography{main}

@article{vaswani2017attention,
  title={Attention is all you need},
  author={Vaswani, Ashish and Shazeer, Noam and Parmar, Niki and Uszkoreit, Jakob and Jones, Llion and Gomez, Aidan N and Kaiser, {\L}ukasz and Polosukhin, Illia},
  journal={Advances in neural information processing systems},
  volume={30},
  year={2017}
}

@article{beyer2024paligemma,
  title={Paligemma: A versatile 3b vlm for transfer},
  author={Beyer, Lucas and Steiner, Andreas and Pinto, Andr{\'e} Susano and Kolesnikov, Alexander and Wang, Xiao and Salz, Daniel and others},
  journal={arXiv preprint arXiv:2407.07726},
  year={2024}
}

@article{openvla,
  author = {Moo Jin Kim and Karl Pertsch and Siddharth Karamcheti and Ted Xiao and Ashwin Balakrishna and Suraj Nair and others},
  title = {{OpenVLA}: An Open-Source Vision-Language-Action Model},
  journal = {arXiv preprint arXiv:2406.09246},
  year = {2024}
}

@article{touvron2023llama,
  title={Llama 2: Open foundation and fine-tuned chat models},
  author={Touvron, Hugo and Martin, Louis and Stone, Kevin and Albert, Peter and Almahairi, Amjad and Babaei, Yasmine and Bashlykov, Nikolay and others},
  journal={arXiv preprint arXiv:2307.09288},
  year={2023}
}

@article{pi0,
  title={{$\pi_0$}: A Vision-Language-Action Flow Model for General Robot Control},
  author={Black, Kevin and Brown, Noah and Driess, Danny and Esmail, Adnan and Equi, Michael and others},
  journal={arXiv preprint arXiv:2410.24164},
  year={2024}
}

@article{pi05,
  title={{$\pi_{0.5}$}: a Vision-Language-Action Model with Open-World Generalization},
  author={Intelligence, Physical and Black, Kevin and Brown, Noah and Darpinian, James and Dhabalia, Karan and Driess, Danny and others},
  journal={arXiv preprint arXiv:2504.16054},
  year={2025}
}

@inproceedings{amin2026pi,
  title={{$\pi_{0.6}$}: A VLA that learns from experience},
  author={Amin, A and Aniceto, R and Balakrishna, A and Black, K and Conley, K and Connors, GB and others},
  booktitle={Proceedings of Robotics: Science and Systems},
  volume={22},
  year={2026}
}

@article{PI07,
  title={{$\pi_{0.7}$}: a Steerable Generalist Robotic Foundation Model with Emergent Capabilities},
  author={Intelligence, Physical and Ai, Bo and Amin, Ali and Aniceto, Raichelle and Balakrishna, Ashwin and Balke, Greg and Black, Kevin and Bokinsky, George and Cao, Shihao and Charbonnier, Thomas and others},
  journal={arXiv preprint arXiv:2604.15483},
  year={2026}
}

@inproceedings{zheng2026x,
  title={X-vla: Soft-prompted transformer as scalable cross-embodiment vision-language-action model},
  author={Zheng, Jinliang and Li, Jianxiong and Wang, Zhihao and Liu, Dongxiu and Kang, Xirui and others},
  booktitle={International Conference on Learning Representations},
  volume={2026},
  pages={60580--60606},
  year={2026}
}

@article{lingbotvla,
  title={A pragmatic vla foundation model},
  author={Wu, Wei and Lu, Fan and Wang, Yunnan and Yang, Shuai and Liu, Shi and others},
  journal={arXiv preprint arXiv:2601.18692},
  year={2026}
}

@article{lingbotvla2,
  title={From Foundation to Application: Improving VLA Models in Practice},
  author={Wu, Wei and Wang, Fangjing and Lu, Fan and Sun, He and Liu, Shi and Wang, Yunnan and others},
  journal={arXiv preprint arXiv:2607.06403},
  year={2026}
}

@article{yang2026abot,
  title={Abot-m0: Vla foundation model for robotic manipulation with action manifold learning},
  author={Yang, Yandan and Zeng, Shuang and Lin, Tong and Chang, Xinyuan and Qi, Dekang and Xiao, Junjin and others},
  journal={arXiv preprint arXiv:2602.11236},
  year={2026}
}

@article{luo2026being,
  title={Being-h0. 5: Scaling human-centric robot learning for cross-embodiment generalization},
  author={Luo, Hao and Wang, Ye and Zhang, Wanpeng and Zheng, Sipeng and Xi, Ziheng and Xu, Chaoyi and Xu, Haiweng and others},
  journal={arXiv preprint arXiv:2601.12993},
  year={2026}
}

@inproceedings{ye2026starvla,
  title={StarVLA-$\alpha$: Reducing complexity in vision-language-action systems},
  author={Ye, Jinhui and Gao, Ning and Yang, Senqiao and Zheng, Jinliang and Wang, Zixuan and others},
  booktitle={European Conference on Computer Vision},
  pages={351--369},
  year={2026},
  organization={Springer}
}

@inproceedings{yu2026twinbrainvla,
  title={Twinbrainvla: Unleashing the potential of generalist vlms for embodied tasks via asymmetric mixture-of-transformers},
  author={Yu, Bin and Lian, Shijie and Lin, Xiaopeng and Shen, Zhaolong and Wu, Changti and Miao, Yuzhuo and Wang, Xinming and others},
  booktitle={Findings of the Association for Computational Linguistics: EMNLP 2026},
  pages={4533--4552},
  year={2026}
}

@article{forcevla,
  title={Forcevla: Enhancing vla models with a force-aware moe for contact-rich manipulation},
  author={Yu, Jiawen and Liu, Hairuo and Yu, Qiaojun and Ren, Jieji and Hao, Ce and Ding, Haitong and others},
  journal={Advances in Neural Information Processing Systems},
  volume={38},
  pages={93409--93439},
  year={2025}
}

@inproceedings{forcevla2,
  title={Forcevla2: Unleashing hybrid force-position control with force awareness for contact-rich manipulation},
  author={Li, Yang and Zhaxizhuoma, Zhaxizhuoma and Jiang, Hongru and Xia, Junjie and Zhang, Hongquan and others},
  booktitle={Proceedings of the IEEE/CVF Conference on Computer Vision and Pattern Recognition},
  pages={8911--8920},
  year={2026}
}

@article{forcedelta,
  title={ForceDelta-VLA: Distilling Force-Conditioned ActionCorrections for Contact-Rich Manipulation},
  author={Dong, Ju and Fu, Yu and Chen, Jian and Liu, Yimeng and Zhao, Haocheng and Zhang, Lei and Bai, Kaixin and others},
  journal={arXiv preprint arXiv:2609.18242},
  year={2026}
}

@article{zhang2026unitacvla,
  title={UniTacVLA: Unified Tactile Understanding and Prediction in Vision Language Action Models},
  author={Zhang, Xidong and Zhang, Yichi and Shi, Jiaxin and Zhu, Fucai and Zhu, Siyu and Wang, Michael Yu and Wu, Xiaojun and Yuan, Weihao},
  journal={arXiv preprint arXiv:2606.31723},
  year={2026}
}

@article{phaforce,
  title = {PhaForce: Phase-Scheduled Visual--Force Policy Learning with Slow Planning and Fast Correction for Contact-Rich Manipulation},
  author = {Wang, Mingxin and Yue, Zhirun and Lu, Renhao and Li, Yizhe and Wang, Zihan and Pan, Guoping and others},
  year = {2026},
  journal={arXiv preprint arXiv:2603.08342},
}

@article{tla,
  title = {Tla: Tactile-language-action model for contact-rich manipulation},
  author = {Peng Hao and Chaofan Zhang and Dingzhe Li and Xiaoge Cao and Xiaoshuai Hao and Shaowei Cui and Shuo Wang},
  year = {2025},
  journal={arXiv preprint arXiv:2503.08548},
}

@article{VTLA,
  title = {VTLA: Vision-tactile-language-action model with preference learning for insertion manipulation},
  author = {Zhang, Chaofan and Hao, Peng and Cao, Xiaoge and Hao, Xiaoshuai and Cui, Shaowei and Wang, Shuo},
  year = {2026},
  journal={Biomimetic Intelligence and Robotics},
}

@inproceedings{atvla,
  title = {AT-VLA: Adaptive Tactile Injection for Enhanced Feedback Reaction in Vision-Language-Action Models},
  author = {Li, Xiaoqi and Cai, Muhe and Xu, Jiadong and Zhu, Juan and Fan, Hongwei and Shen, Yan and others},
  booktitle={Biomimetic Intelligence and Robotics},
  pages = {28764--28774},
  year = {2026}
}

@article{tactilevla,
  title   = {Tactile-VLA: Unlocking Vision-Language-Action Model's Physical Knowledge for Tactile Generalization},
  author  = {Huang, Jialei and Wang, Shuo and Lin, Fanqi and Hu, Yihang and others},
  journal = {arXiv preprint arXiv:2507.09160},
  year    = {2025},
}

@article{Qwen2,
  title   = {Qwen2-VL: Enhancing Vision-Language Model's Perception of the World at Any Resolution},
  author  = {Wang, Peng and Bai, Shuai and Tan, Sinan and Wang, Shijie and Fan, Zhihao and Bai, Jinze and others},
  journal = {arXiv preprint arXiv:2409.12191},
  year    = {2024},
}

@article{team2024gemma,
  title={Gemma: Open models based on gemini research and technology},
  author={Team, Gemma and Mesnard, Thomas and Hardin, Cassidy and Dadashi, Robert and Bhupatiraju, Surya and others},
  journal={arXiv preprint arXiv:2403.08295},
  year={2024}
}

@article{lou2026dream,
  title={Dream-Tac: A Unified Tactile World Action Model for Contact-Rich Robot Manipulation},
  author={Lou, Yunfan and Ye, Yifan and Fu, Yankai and Cen, Jun and Chi, Xiaowei and Lyu, Yaoxu and others},
  journal={arXiv preprint arXiv:2606.08737},
  year={2026}
}
	}
\end{document}